\pdfoutput=1
\documentclass[letterpaper,10pt,conference]{ieeeconf}

\IEEEoverridecommandlockouts
\usepackage{amsmath}
\usepackage{amssymb}
\usepackage{booktabs}
\usepackage{graphicx}
\usepackage{url}
\usepackage[hidelinks]{hyperref}
\usepackage{xcolor}
\usepackage{placeins}
\usepackage{capt-of}
\newcommand{\method}{\textsc{GestureFuse}}

\title{GestureFuse: Exploiting Graph--Transformer Complementarity\\ for Traffic-Control Gesture Recognition}

\author{Cheng~Li$^{1,*,\dagger}$, Renjun~Gao$^{2,*,\dagger}$, Chenhao~Guan$^{2,*}$,
Boyi~Fu$^{3}$, and Xiaolei~Ren$^{2}$\\[0.35em]
{\small $^{1}$The Hong Kong University of Science and Technology}\\
{\small $^{2}$Macau University of Science and Technology}\quad
{\small $^{3}$The Hong Kong University of Science and Technology (Guangzhou)}\\[0.15em]
{\footnotesize $^{*}$Equal contribution\qquad $^{\dagger}$Corresponding authors:
\href{mailto:clieo@connect.ust.hk}{clieo@connect.ust.hk},
\href{mailto:renjgao@gmail.com}{renjgao@gmail.com}}
}
\begin{document}

\IEEEaftertitletext{%
  \begin{minipage}{\textwidth}
    \centering
    \includegraphics[width=0.96\linewidth]{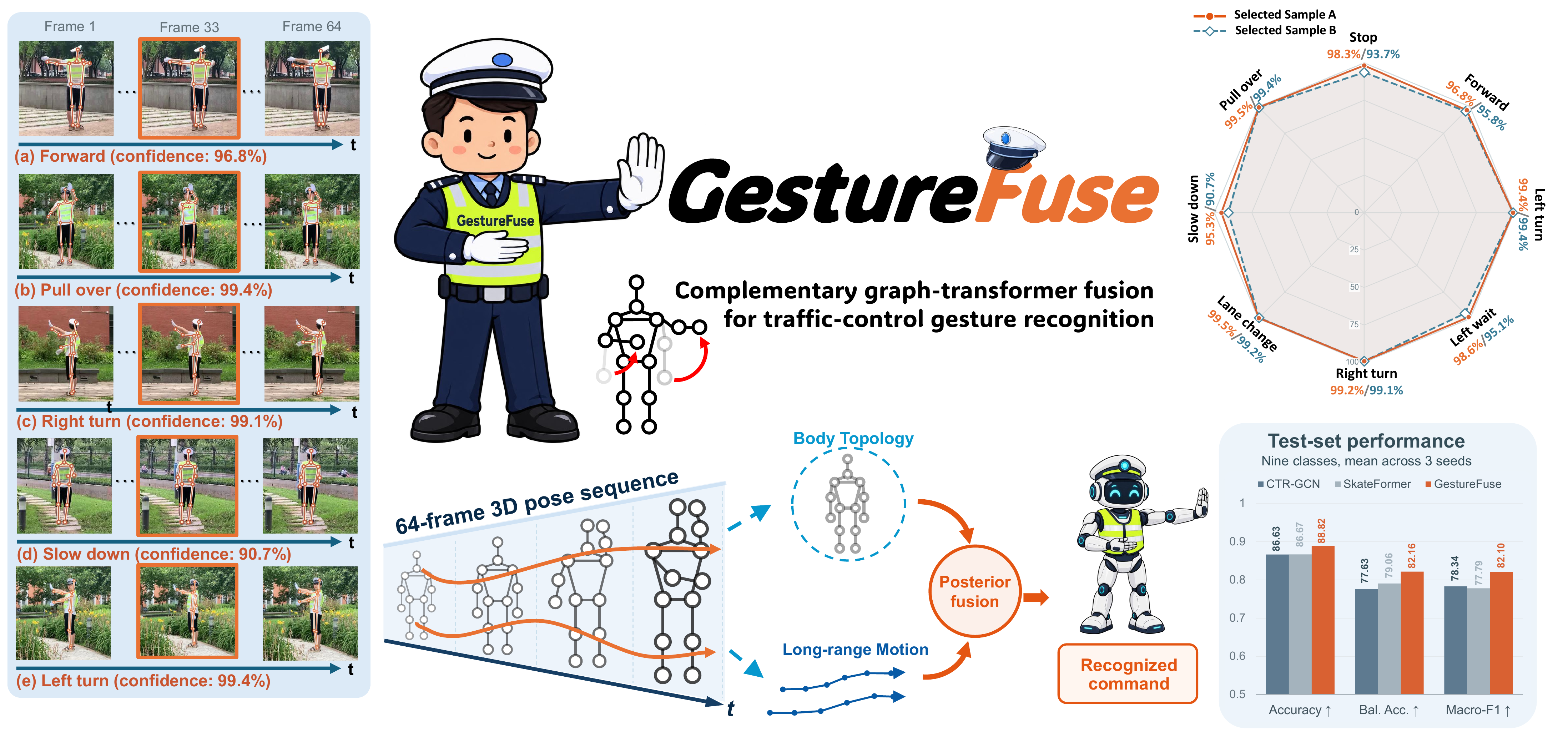}
    \captionof{figure}{Facing brief transitions and similar arm configurations in traffic-control gestures, we propose \method, a complementary graph--transformer framework for traffic-command recognition. Graph and transformer encoders independently process the same centered 64-frame 3D pose sequence, emphasizing topology-aware joint relations and long-range skeletal--temporal dependencies, respectively. Validation-selected posterior fusion combines their class posteriors to produce the final command prediction. \method\ achieves 88.82\% accuracy and 82.10\% Macro-F1 over three seeds on the nine-class CTPGesture v2 protocol, outperforming both individual encoders.}
    \label{fig:teaser}
  \end{minipage}
  \vspace{0.6em}
}

\maketitle
\thispagestyle{empty}
\pagestyle{empty}

\begin{abstract}
Reliable recognition of human traffic-control gestures is essential for autonomous 
vehicles operating when fixed traffic infrastructure is unavailable or temporarily 
overridden. Brief commands, frequent transitions, and similar arm configurations require 
joint modeling of articulated pose and temporal context. Graph networks and
skeletal transformers offer complementary inductive biases for modeling these
cues, emphasizing joint coordination and long-range motion dependencies,
respectively. We present \method, a graph--transformer framework that combines
the class posteriors of two independently trained encoders operating on the
same pose sequence. A single fusion coefficient is selected on source-disjoint
validation videos and shared across all samples and classes. Pairwise margin
analysis characterizes how this mixture resolves branch disagreements and
relates prediction changes to the evidence supplied by each encoder. Across
three seeds on the nine-class CTPGesture v2 protocol, \method\ achieves
$88.82\%$ accuracy and $82.10\%$ Macro-F1, improving on the strongest individual
branch for each metric by $2.15$ and approximately $3.8$ percentage points,
respectively.
Recording-level analyses and targeted ablations further demonstrate the contribution
of complementary graph- and transformer-based skeletal evidence. Together, these results 
establish posterior-level graph-transformer fusion as an effective approach to 
class-balanced traffic-command recognition in human-directed traffic-control scenarios.
\mbox{Visit \href{https://gesturefuse.renjun.one/}{\textcolor[HTML]{0066FF}{the project webpage}} for details.}
\end{abstract}

\section{Introduction}
\label{sec:intro}

Autonomous vehicles must interpret temporary instructions from human traffic
controllers as well as traffic lights and road signs. During signal failures,
road construction, traffic incidents, and emergency response, these gestures
communicate the intended movement of surrounding vehicles and may take
precedence over fixed signals. Recognizing this visual communication is
therefore an important component of automated driving in manually controlled
environments \cite{wiederer2020tcg}. Compared with trimmed action recognition,
traffic-gesture recognition requires distinguishing brief commands within
continuous recordings containing long inactive intervals and frequent
transitions. Officers may occupy a small region of a cluttered scene, and
viewpoint, scale, and appearance vary across recordings. Moreover, distinct
commands can share similar instantaneous arm configurations. Accurate
recognition consequently depends on both arm--torso geometry and its evolution
over time. Pose sequences retain this articulated structure while reducing
dependence on clothing and background appearance
\cite{xiong2021mdgcn,liu2021trafficgcn,zhang2022hlp,holzbock2022stmlp}.

Prior traffic-gesture methods have investigated graph connectivity
\cite{xiong2021mdgcn,liu2021trafficgcn}, body-part partitioning
\cite{zhang2022hlp}, spatio-temporal multilayer perceptrons
\cite{holzbock2022stmlp}, recurrent--transformer models
\cite{zhang2026bilstmtransformer}, and position or orientation priors
\cite{he2026prgraph}. More broadly, skeleton-based recognition has evolved along 
complementary directions for structured spatial reasoning and long-range temporal modeling. 
Graph networks emphasize structured joint relations \cite{yan2018stgcn,chen2021ctrgcn}, whereas
skeletal attention models capture dependencies across joints and frames
\cite{shi2020dstanet,do2024skateformer}. Hybrid architectures such as MG-GCT combine graph convolution and transformer modeling within a jointly trained network for traffic gestures \cite{guo2024mggct}. This leaves unresolved whether independently optimized graph and transformer recognizers provide complementary class evidence when operating on identical pose observations, and whether such complementarity can be exploited at the decision level.

We address this question with \method, a controlled shared-input graph–transformer framework that combines CTR-GCN \cite{chen2021ctrgcn} and SkateFormer
\cite{do2024skateformer}. The two branches independently process the same
64-frame, 49-joint pose sequence, emphasizing topology-aware joint relations
and long-range skeletal--temporal dependencies, respectively. Their aligned
class posteriors are combined using one coefficient selected on
source-disjoint validation videos and fixed for testing. Sharing the coefficient across samples and classes reduces fusion selection to a one-dimensional validation problem and adds only $O(|\mathcal C|)$ computation per prediction. Pairwise margin analysis
exposes how the coefficient changes class orderings when the branches
disagree. The resulting formulation provides an explicit link between branch 
evidence and the fused decision while preserving independent evaluation of each encoder.

On the nine-class CTPGesture v2 protocol, \method\ achieves $88.82\%$ accuracy
and $82.10\%$ Macro-F1 across three seeds. These results exceed SkateFormer in
accuracy by $2.15$ percentage points and CTR-GCN in Macro-F1 by
approximately $3.8$ percentage points.

The main contributions are as follows:
\begin{itemize}
  \item \textbf{Shared-input graph--transformer recognition.}
  We develop a dual-encoder framework that combines topology-aware graph reasoning and long-range skeletal--temporal attention over identical pose sequences. Independent optimization preserves the architecture-specific modeling characteristics of the two encoders and supports direct evaluation of each branch's contribution.

  \item \textbf{Validation-constrained scalar posterior fusion.}
  We formulate a convex mixture of aligned class posteriors controlled by one coefficient shared across samples and classes. The coefficient is selected on source-disjoint validation videos using recognition and calibration criteria and then fixed for test evaluation.

  \begin{samepage}
  \item \textbf{Pairwise characterization of fusion decisions.}
  We derive class-pair margin thresholds that characterize how the global coefficient changes class orderings when the branches disagree. The analysis identifies when fusion can recover the correct ordering under opposing branch
  margins and when a shared top-ranked error cannot be corrected by any admissible coefficient.
  \par\end{samepage}
\end{itemize}

\section{Related Work}
\label{sec:related}

\subsection{Traffic-Gesture Recognition from Pose Sequences}

Traffic-gesture recognition identifies a compact command vocabulary within continuous sequences containing inactive periods and transitional motion. Wiederer et al. introduced the TCG benchmark for autonomous-driving gesture
recognition \cite{wiederer2020tcg}. Subsequent traffic-officer recognition
methods investigated skeleton extraction with multichannel dilated graphs
\cite{xiong2021mdgcn}, graph convolution in natural scenes
\cite{liu2021trafficgcn}, height-layer body partitions \cite{zhang2022hlp}, and
spatio-temporal multilayer perceptrons \cite{holzbock2022stmlp}. These studies
establish articulated pose as a useful representation for traffic commands and
highlight the importance of temporal context and body structure.

Recent work includes recurrent--transformer modeling
\cite{zhang2026bilstmtransformer}, position--rotation graph structure
\cite{he2026prgraph}, and pedestrian gesture analysis
\cite{mahdi2026gesturematters}. PR-Graph introduces CTPGesture v2 with
command and orientation annotations, supporting a 33-class joint formulation
\cite{he2026prgraph}. RSC-GestureNet emphasizes reliable causal recognition
and selective command emission \cite{rscgesturenet2026repo}. We adopt a nine-class 
center-frame recognition protocol and reserve orientation labels for subgroup analysis.
For context, Table~\ref{tab:main} separately reports a dense RSC-GestureNet adaptation 
under a 33-class validation-transfer setting.

\subsection{Graph and Attention Models for Skeleton Dynamics}

Skeleton-based recognition models human motion as a structured time series.
ST-GCN established spatio-temporal message passing over joints and frames
\cite{yan2018stgcn}; AS-GCN \cite{li2019asgcn} and two-stream adaptive GCNs
\cite{shi2019twostream} extended this formulation with learned joint
relations. Shift-GCN \cite{cheng2020shiftgcn}, MS-G3D \cite{liu2020msg3d},
MST-GCN \cite{chen2021mstgcn}, and InfoGCN \cite{chi2022infogcn} further
improved feature exchange, multi-scale modeling, and representation learning.
CTR-GCN refines topology by channel, adapting joint relations to the observed
motion \cite{chen2021ctrgcn}.

Attention-based models offer complementary mechanisms for aggregating
skeletal information. DSTA-Net decouples spatial and temporal attention
\cite{shi2020dstanet}, while SkateFormer partitions joints and frames to
model skeletal--temporal dependencies efficiently \cite{do2024skateformer}.
Other approaches investigate hierarchical graphs \cite{lee2023hdgcn}, pose
heatmap volumes \cite{duan2022posec3d}, recognition in unconstrained scenes
\cite{duan2023skeletr}, and motion pretraining \cite{zhu2023motionbert}.
\method\ exploits the distinct representations learned by graph and attention encoders 
while combining their predictions over the same observed motion. This
shared-input formulation makes their contributions directly comparable at the
level of individual gesture sequences.

\subsection{Complementary Graph--Transformer Fusion}

Graph--transformer integration has been explored within traffic-gesture
architectures, including the motion-guided MG-GCT \cite{guo2024mggct}.
\method\ instead combines the outputs of independently optimized encoders.
Decision-level fusion allows such predictors to retain their representations
while contributing complementary class evidence. Deep ensembles demonstrate
the value of combining independent predictors for recognition and uncertainty
estimation \cite{lakshminarayanan2017ensembles}. Temperature scaling
\cite{guo2017calibration} and Dirichlet calibration \cite{kull2019dirichlet}
address a related issue by adjusting predicted probabilities. Here, a convex
posterior mixture provides a direct means of integrating graph-local and
long-range skeletal evidence with one coefficient.

Coefficient selection is particularly important for imbalanced gesture data
with correlated windows. A global weight keeps the selection dimension small,
while source-disjoint validation separates the recordings used for coefficient
selection from those used for evaluation. Calibration assessment also depends
on the estimator and sampling procedure
\cite{widmann2019calibration,popordanoska2022calibration,kumar2019verified};
calibration-oriented training provides an additional means of shaping
confidence \cite{mukhoti2020focal}. We therefore select the mixture using
recognition and calibration criteria together, then examine uncertainty through
recording-level resampling and complementary probability-based metrics.

\begin{figure*}[t]
  \centering
  \ifdefined\GestureFusePublicSafe
    \includegraphics[width=\textwidth,height=0.299\textwidth,keepaspectratio]{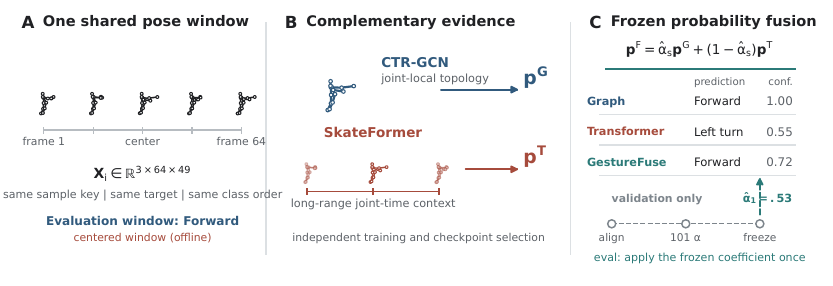}
  \else
    \includegraphics[width=0.97\textwidth]{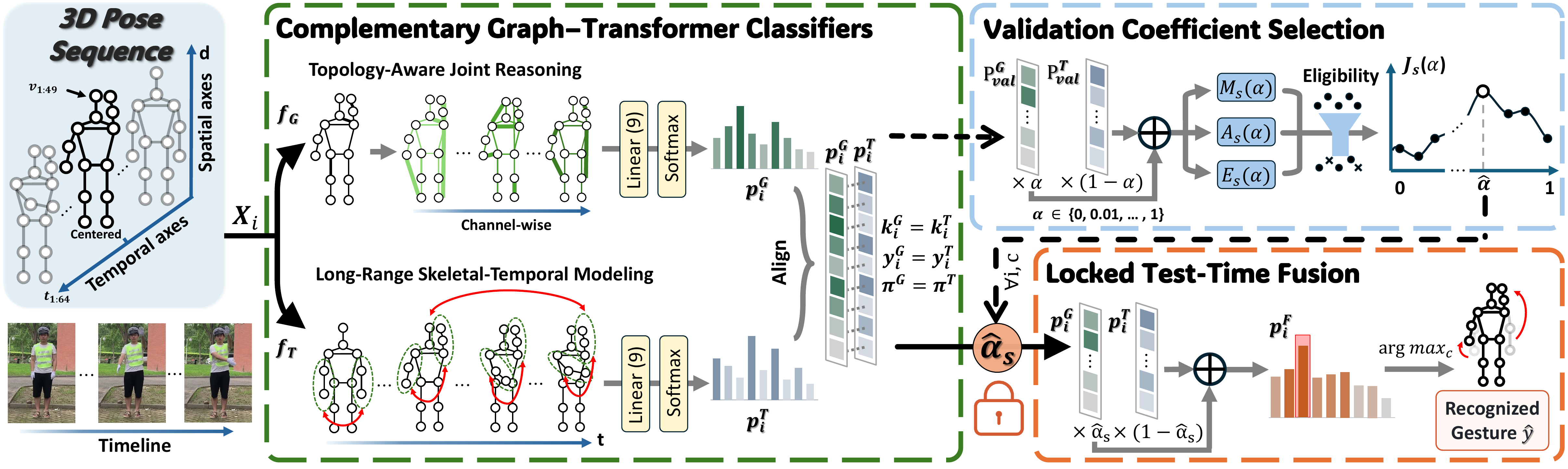}
  \fi
  \caption{Overview of \method. CTR-GCN and SkateFormer independently process
  a 64-frame, 49-joint 3D pose sequence centered on the target frame and produce
  nine-class posteriors. Sample identifiers, target labels, and class order
  are checked for consistency in the evaluation records. Recognition and
  calibration criteria on source-disjoint validation videos determine one scalar coefficient per
  seed, shared across samples and classes. The selected coefficient is fixed
  and applied to the corresponding test predictions. }
  \label{fig:framework}
\end{figure*}

\section{Methodology}
\label{sec:method}
\subsection{Data Processing}

We use a nine-class center-label protocol on CTPGesture v2
\cite{he2026prgraph}, comprising an inactive state and eight traffic-control
commands. The source-video partition contains 8 training, 2 validation, and
5 test recordings, yielding 2,169, 476, and 1,965 windows, respectively.
Overlapping windows from each recording are assigned to the same partition.
The test set contains 1,219 inactive windows and 51--113 windows per active
class, motivating class-balanced evaluation alongside overall accuracy.
Each sample retains its source identifier, temporal extent, label, and
skeleton specification. Prediction records additionally store the ordered
class probabilities, predicted label, and confidence, enabling paired analysis
of the same observations across methods and source recordings.

\subsection{Task and Pose Representation}

Let $\mathcal C=\{0,\ldots,8\}$ denote the inactive state and eight traffic
commands. Each sample contains 64 consecutive frames and is assigned the
label of its center frame. The preprocessed pose tensor is
$\mathbf X_i\in\mathbb R^{3\times64\times49\times1}$, whose dimensions
index coordinates, frames, joints, and the tracked person. Each recognizer
produces a distribution in the nine-class simplex $\Delta^8$. The classifiers
use skeletal coordinates as their sole input, while orientation and source
identifiers serve as evaluation metadata.

The 49-joint ordering follows the common VIBE/SPIN--OpenPose export
\cite{cao2017openpose,kolotouros2019spin,kocabas2020vibe}. For coordinate
channel $d$, frame $t$, and joint $v$, preprocessing applies
\begin{equation}
 \bar{x}_{i,d,t,v}=m_{i,t}\bigl(x_{i,d,t,v}-x_{i,d,t,0}\bigr),
 \label{eq:center}
\end{equation}
where $m_{i,t}$ is one for an available frame and zero otherwise. We term this
operation \emph{fixed-reference centering}, with joint 0 serving as the
reference in the exported joint ordering.
The centered coordinates form $\mathbf X_i$. Pose-estimation quality can
influence recognition \cite{sun2019hrnet}; accordingly, pose extraction,
missing-frame handling, window endpoints, and joint and class orderings are
held fixed across the primary comparison.

\subsection{Complementary Skeleton Encoders}

Two independently trained classifiers process the shared pose tensor.
The graph branch $f_G$ adapts CTR-GCN to refine channel-specific relations
over the skeleton \cite{chen2021ctrgcn}. The attention branch $f_T$ adapts a
four-stage SkateFormer with depths $(2,2,2,2)$ and channel widths
$(96,192,192,192)$ to the 49-joint input \cite{do2024skateformer}. Their logits and posteriors are
\begin{equation}
 \mathbf z_i^b=f_b(\mathbf X_i;\theta_b),\qquad
 \mathbf p_i^b=\operatorname{softmax}(\mathbf z_i^b),\quad b\in\{G,T\}.
 \label{eq:branches}
\end{equation}
CTR-GCN is optimized with cross-entropy. SkateFormer uses label smoothing
with $\epsilon=0.05$ over $N$ training samples:
\begin{equation}
 \mathcal L_T=-\frac{1}{N}\sum_{i,c}
 \left[(1-\epsilon)\mathbf 1(y_i=c)+\frac{\epsilon}{|\mathcal C|}\right]
 \log p^T_{i,c}.
 \label{eq:loss}
\end{equation}

The two encoders derive complementary representations from the same skeletal observation.
CTR-GCN propagates features through channel-refined joint relations,
emphasizing articulated coordination across the sequence. SkateFormer
partitions joints and frames for attention, allowing temporally separated
motion phases to exchange information within the window. The two encoders are optimized independently to learn architecture-specific skeletal representations. Once training is complete and their checkpoints are fixed, posterior-level fusion combines their class posteriors without updating either encoder. Standalone and fused evaluations use predictions from the same fixed checkpoints, isolating the effect of posterior fusion from changes in representation learning.

\subsection{Aligned Graph--Transformer Fusion}

Each prediction is associated with a sample key $k_i$ comprising the source
video, temporal extent, partition, and row identifier. The prediction record
stores this key, the target $y_i$, the ordered class map $\pi$, and all nine
class probabilities. For prediction sets $P^G$ and $P^T$, alignment requires
\begin{equation}
 \operatorname{Align}(P^G,P^T)=
 \bigwedge_i[k_i^G=k_i^T\land y_i^G=y_i^T]\land(\pi^G=\pi^T).
 \label{eq:align}
\end{equation}
Fusion is applied only to one-to-one matched prediction records with consistent sample identities and class ordering, including unique sample keys, matched row counts, 
finite values, and valid probability vectors. These checks establish correspondence 
between evaluation records. Inference aligns sample keys and class order, while target
labels are used for evaluation.

For seed $s$, \method\ combines class probabilities as
\begin{align}
 \mathbf q_i^s(\alpha)
 &=\alpha\mathbf p_i^{G,s}+(1-\alpha)\mathbf p_i^{T,s},\label{eq:fuse}\\
 \mathbf p_i^{F,s}(\alpha)
 &=\frac{\operatorname{clip}(\mathbf q_i^s,10^{-12},1)}
 {\sum_c\operatorname{clip}(q_{i,c}^s,10^{-12},1)}.
 \label{eq:renorm}
\end{align}
Clipping and renormalization maintain numerical stability. For each seed $s$,
a single coefficient $\alpha_s\in[0,1]$ weights the graph posterior, while
$1-\alpha_s$ weights the transformer posterior. The coefficient is shared
across all samples and classes.

\subsection{Pairwise Decision Analysis}

The convex mixture provides an explicit relationship between branch evidence
and the fused decision. Before numerical clipping, each fused class score
lies between the corresponding branch scores. Fusion changes a prediction by
reordering these scores as their relative contributions vary. This property
is particularly useful for interpreting sequences on which one encoder
identifies the correct command while the other favors a competing class.
The examples in Fig.~\ref{fig:qualitative} illustrate both directions of this
complementarity together with a case of consistent cross-branch agreement.

For candidate classes $a$ and $b$, define the branch margins as
$\delta_i^G=p_{i,a}^G-p_{i,b}^G$ and
$\delta_i^T=p_{i,a}^T-p_{i,b}^T$. The fused margin is
\begin{equation}
 \Delta_i^{a,b}(\alpha)=\delta_i^T+
 \alpha(\delta_i^G-\delta_i^T).
 \label{eq:margin}
\end{equation}
Margins with the same sign preserve their pairwise ordering throughout
$\alpha\in[0,1]$. When the signs differ, the ordering changes at
$\alpha_i^*=-\delta_i^T/(\delta_i^G-\delta_i^T)$.
Because these thresholds vary across sequences, a single coefficient can
preserve the graph decision for some observations and the transformer
decision for others. Across class pairs, the thresholds partition the
coefficient range into intervals with fixed class orderings, apart from ties.
The global mixture therefore accommodates different disagreement patterns
through the branch margins themselves.

The same analysis characterizes shared prediction errors. If both encoders
strictly rank the same incorrect class first, the convex mixture retains that
class for every admissible coefficient. In contrast, opposing margins allow
fusion to recover the correct ordering when the selected weight lies on the
appropriate side of the crossing. These properties explain the role of the
scalar parameter and motivate its use with a small validation set. Classwise
weighting would introduce nine coefficients, whereas the common weight has
one degree of freedom for 476 validation windows from two recordings.
The selected mixture thus combines interpretable decision behavior with a
compact validation problem.

\subsection{Validation-Based Coefficient Selection}

We search $\mathcal G=\{0,0.01,\ldots,1\}$ using aligned validation
predictions. Let $M_s(\alpha)$, $A_s(\alpha)$, and $E_s(\alpha)$ denote
Macro-F1, balanced accuracy, and expected calibration error (ECE),
respectively, expressed on the $[0,1]$ scale. The selection objective is
\begin{equation}
 J_s(\alpha)=M_s(\alpha)+0.25A_s(\alpha)-0.10E_s(\alpha).
 \label{eq:objective}
\end{equation}
Candidate coefficients must also satisfy recognition and calibration
constraints relative to the branch results:
\begin{align}
M_s(\alpha)&\geq\max(M_s^G,M_s^T)-0.002,\nonumber\\
A_s(\alpha)&\geq\max(A_s^G,A_s^T)-0.005,\nonumber\\
E_s(\alpha)&\leq\max(E_s^G,E_s^T)+0.010.
\label{eq:feasible}
\end{align}
The recognition bounds use the stronger validation branch for each
metric. The ECE bound permits at most $0.010$ above the larger branch ECE. All three seeds have
nonempty feasible sets. Feasible candidates are ranked lexicographically by
\begin{equation}
 \bigl(J_s,M_s,A_s,-E_s,-|\alpha-0.5|\bigr),
 \label{eq:tiebreak}
\end{equation}
which prioritizes the objective, then the individual metrics and proximity
to equal weighting.

The selected graph weights are $0.53$, $0.78$, and $0.46$ for three seeds. Each $\hat\alpha_s$ is fixed after validation and applied to the corresponding test predictions. Standalone and fused results are computed from the same validation-selected branch checkpoints. Performance is summarized across the three seed-specific mixtures without averaging posteriors across seeds.

The implementation separates model selection, posterior export, and
evaluation. Checkpoints selected by validation Macro-F1 produce aligned
validation predictions. The selector evaluates the 101 coefficient candidates
on these predictions and records their feasibility, ranking tuples, and
selected coefficient. The encoder checkpoints and coefficient are then fixed
for test evaluation. Prediction records, configuration metadata, and checksums
support reproducible evaluation from the saved outputs. The standalone branch
results and fused results use the same validation-selected checkpoints.

\subsection{Efficient Posterior Fusion}

Given the branch posteriors, scalar weighting, clipping, and normalization
require $O(|\mathcal C|)$ additional computation and memory. The fusion layer
contains one selected coefficient and leaves the encoder architectures
unchanged. Its cost is therefore small relative to computing both branch
outputs, which remains the principal inference cost of the dual-encoder
framework.

The centered 64-frame window provides observations before and after the target frame, supporting the distinction between commands with similar instantaneous poses but different motion trajectories. Both branches process this identical temporal context. Because the window includes future observations, the present formulation addresses offline center-frame recognition rather than causal online inference.

\section{Experiments}
\label{sec:experiments}

\subsection{Experimental Setup}
\label{subsec:setup}

The primary comparison evaluates CTR-GCN, SkateFormer, and \method\ using
identical 49-joint pose tensors, sample keys, temporal windows, center-frame
labels, source-video partitions, and metrics. VIBE-GRU provides a lower-capacity framewise reference,
combining VIBE-derived pose features~\cite{kocabas2020vibe}
with a GRU gesture classifier implemented in this study. The dense RSC-GestureNet adaptation
\cite{rscgesturenet2026repo} uses COCO17-style poses and is reported separately
under a 33-class command--orientation validation protocol.

\begin{figure*}[t]
  \centering
  \includegraphics[width=0.94\textwidth]{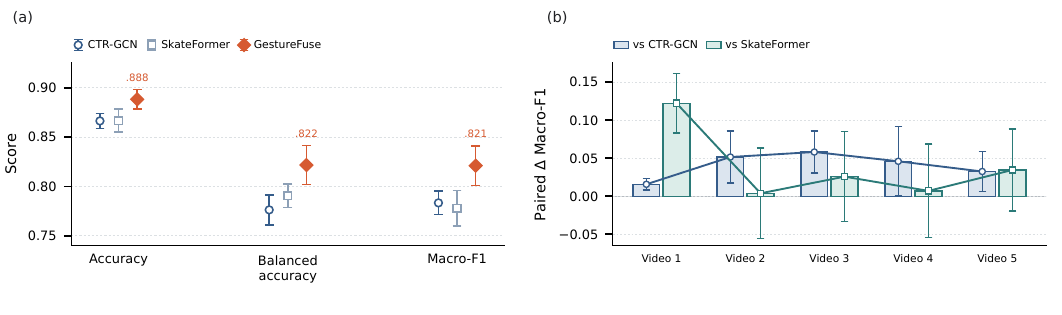}
  \caption{Recognition performance under shared evaluation inputs.
  (a) Mean and standard deviation across three seeds for the primary methods.
  (b) Paired Macro-F1 gains of \method\ over each branch on all five test
  recordings. Bars show seed-averaged gains, error bars show one standard
  deviation, and lines connect the recording-level means.}
  \label{fig:results}
\end{figure*}

\begin{table*}[t]
  \centering
  \small
  \caption{Three-seed results (mean $\pm$ sample standard deviation). The
  primary nine-class comparison uses shared center-label test inputs and
  validation-selected checkpoints; VIBE-GRU is a framewise reference. The
  auxiliary 33-class validation-transfer result appears in a separate block.
  Bold identifies the best value in the primary comparison.}
  \label{tab:main}
  \small
  \setlength{\tabcolsep}{3.6pt}
  \begin{tabular}{lccccc}
    \toprule
    Method & Accuracy $\uparrow$ & Bal. Acc. $\uparrow$ & Macro-F1 $\uparrow$
    & Weighted-F1 $\uparrow$ & ECE $\downarrow$ \\
    \midrule
    \multicolumn{6}{l}{\textit{Nine-class test evaluation}} \\
    VIBE-GRU (frame ref.)~\cite{kocabas2020vibe}
    & $0.7414{\pm}0.0075$ & $0.7220{\pm}0.0178$
    & $0.6221{\pm}0.0234$ & $0.7586{\pm}0.0097$
    & $0.0508{\pm}0.0117$ \\
    CTR-GCN~\cite{chen2021ctrgcn}
    & $0.8663{\pm}0.0076$ & $0.7763{\pm}0.0151$
    & $0.7834{\pm}0.0118$ & $0.8655{\pm}0.0083$
    & $0.0720{\pm}0.0032$ \\
    SkateFormer~\cite{do2024skateformer}
    & $0.8667{\pm}0.0117$ & $0.7906{\pm}0.0118$
    & $0.7779{\pm}0.0178$ & $0.8666{\pm}0.0111$
    & $0.0418{\pm}0.0032$ \\
    \textbf{\method\ (proposed)}
    & $\mathbf{0.8882{\pm}0.0100}$ & $\mathbf{0.8216{\pm}0.0198}$
    & $\mathbf{0.8210{\pm}0.0200}$ & $\mathbf{0.8881{\pm}0.0099}$
    & $\mathbf{0.0408{\pm}0.0124}$ \\
    \midrule
    \multicolumn{6}{l}{\textit{Auxiliary 33-class validation transfer}} \\
    RSC-GestureNet (transfer)~\cite{rscgesturenet2026repo}
    & $0.7371{\pm}0.0168$ & $0.7446{\pm}0.0313$
    & $0.6387{\pm}0.0149$ & $0.7494{\pm}0.0089$
    & $0.2425{\pm}0.0174$ \\
    \bottomrule
  \end{tabular}
\end{table*}

CTR-GCN, SkateFormer, and VIBE-GRU are trained for up to 80 epochs over three
seeds, using architecture-specific optimization settings. CTR-GCN uses Adam with
learning rate $5\times10^{-4}$ and batch size 64. SkateFormer has 3.52M
parameters, uses a head configuration of $H=32$, and is trained with AdamW at learning
rate $4\times10^{-4}$ and batch size 32. VIBE-GRU uses AdamW with a
96-dimensional hidden state. For each run, the checkpoint with the highest
validation Macro-F1 is retained. Fusion coefficients are selected from the
validation predictions, and all checkpoints and coefficients are fixed before
test evaluation. \method\ combines the corresponding same-seed branch
predictions.

We report accuracy, balanced accuracy, Macro-F1, weighted-F1, and 15-bin ECE.
Metrics are computed separately for each seed and
reported as mean $\pm$ sample standard deviation. Paired percentile intervals
for \method-minus-baseline differences are obtained by resampling the five
test recordings 20,000 times \cite{efron1993bootstrap}, retaining all windows
within each sampled recording.

\begin{figure*}[t]
  \centering
  \ifdefined\GestureFusePublicSafe
    \includegraphics[width=0.91\textwidth,height=0.341\textwidth,keepaspectratio]{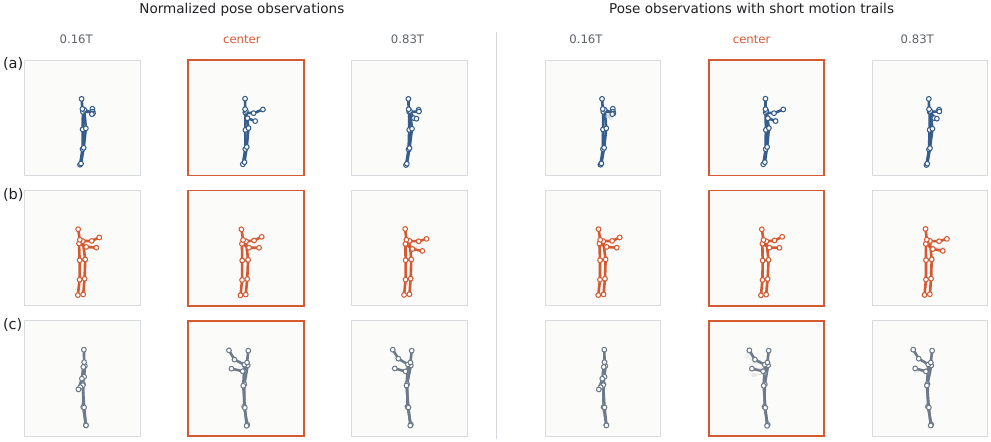}
    \caption{Pose sequences illustrating complementary and shared branch
    errors. Each row shows normalized poses sampled from one 64-frame
    evaluation window. The corresponding seed-1 predictions are reported in
    Table~\ref{tab:qualitative}.}
  \else
    \includegraphics[width=0.85\textwidth]{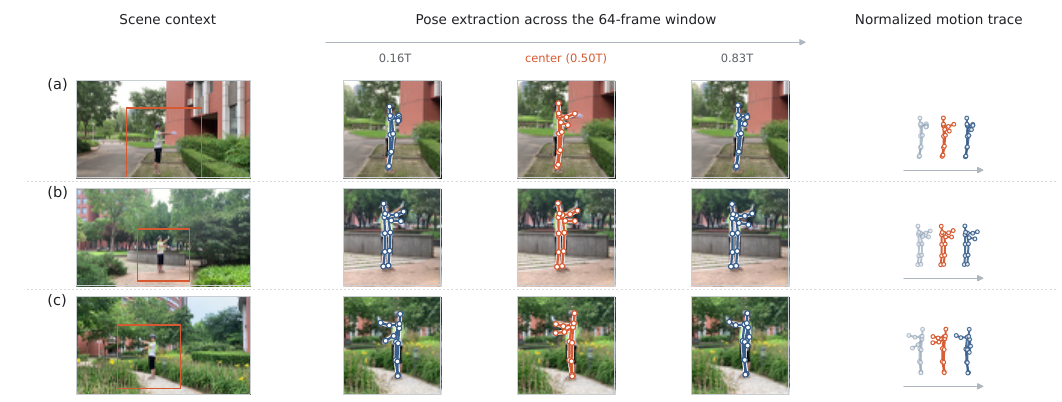}
    \caption{Qualitative visualization of CTPGesture v2 examples. Each row shows
an RGB frame, the corresponding officer crop with the extracted
VIBE/SPIN--OpenPose skeleton, and normalized poses sampled at $0.16T$, the
center frame, and $0.83T$ from the 64-frame input sequence ($T=64$).}
  \fi
  \label{fig:qualitative}
\end{figure*}

\begin{table*}[t]
  \centering
  \caption{Seed-1 predictions for the sequences in
  Fig.~\ref{fig:qualitative}. Parentheses give top-1 confidence rounded to two
  decimals. All predictions refer to the same observations and class ordering.}
  \label{tab:qualitative}
  \small
  \setlength{\tabcolsep}{7.0pt}
  \begin{tabular}{ccccc}
    \toprule
    Panel & Ground truth & CTR-GCN & SkateFormer & \method \\
    \midrule
    (a) & Forward & Forward (1.00) & Left turn (0.55) & Forward (0.72) \\
    (b) & Pull over & Right turn (0.51) & Pull over (0.98) & Pull over (0.71) \\
    (c) & Left turn & Left turn (1.00) & Left turn (0.99) & Left turn (0.99) \\
    \bottomrule
  \end{tabular}
\end{table*}

\subsection{Main Recognition Results}

Table~\ref{tab:main} summarizes the recognition results. \method\ achieves
$0.8882$ accuracy, $0.8216$ balanced accuracy, $0.8210$ Macro-F1, and $0.8881$
weighted-F1, with ECE of $0.0408$. It has the highest mean recognition scores
and the lowest mean ECE among the three methods in the primary comparison,
and also exceeds the VIBE-GRU reference in recognition performance. Relative
to CTR-GCN and SkateFormer, the accuracy gains are $2.19$ and $2.15$ percentage
points, respectively. The improvement extends to class-balanced metrics,
indicating benefits across the command vocabulary as well as under the
observed class distribution.

The branch results reveal different strengths. SkateFormer has higher
balanced accuracy than CTR-GCN ($0.7906$ versus $0.7763$), while CTR-GCN has
higher Macro-F1 ($0.7834$ versus $0.7779$). Their overall accuracies are
nearly identical. Combining the posteriors raises both class-balanced scores
above those of either encoder, consistent with complementary precision and
recall profiles. Weighted-F1 also increases from $0.8655$ and $0.8666$ to
$0.8881$, showing that these gains coexist with improved performance under
the empirical class frequencies. The ECE reduction is substantial relative
to CTR-GCN ($0.0720$ to $0.0408$) and small relative to SkateFormer
($0.0418$ to $0.0408$).

Expressed relative to the remaining classification error, the accuracy gains
correspond to reductions of approximately $16.4\%$ over CTR-GCN and $16.1\%$
over SkateFormer. These ratios describe the effect size under the evaluated
class distribution. Balanced accuracy and Macro-F1 provide complementary
views by weighting classes equally: the former averages recall, while the
latter also incorporates precision. Their joint improvement is important
because inactive windows account for most test observations. An increase in
accuracy alone could obscure errors on the less frequent active commands.

The dense RSC-GestureNet adaptation yields $0.7371$ accuracy, $0.7446$
balanced accuracy, $0.6387$ Macro-F1, and $0.7494$ weighted-F1 in its 33-class
validation-transfer setting. Its ECE of $0.2425$ is computed from the saved
33-way probabilities using the same 15-bin estimator. This reference
characterizes transfer to the joint command--orientation task. Its distinct
label space and evaluation partition are indicated separately in the table;
the primary ranking and paired statistical analysis concern the common
nine-class test protocol.

Figure~\ref{fig:results}(a) presents the mean recognition scores and seed
variation for the primary comparison. Panel (b) resolves the Macro-F1
improvement by source recording. Seed-averaged gains are positive on all
five videos against both branches, although their magnitude varies by
recording. The same coefficient within each seed is applied to every video
and gesture class. The recording-level pattern therefore shows that the
benefit of combining branch evidence extends across the evaluated sources,
rather than being confined to a single recording. The following analysis
quantifies uncertainty at this source level.

\subsection{Source-Level Statistical Analysis}

Recording-level resampling complements the per-video results in
Fig.~\ref{fig:results}(b). Against CTR-GCN, the paired intervals for accuracy,
balanced accuracy, and Macro-F1 are all above zero. Against SkateFormer, the
accuracy interval is $[0.0009,0.0418]$ and the Macro-F1 interval is
$[0.0095,0.0815]$. The balanced-accuracy interval, $[-0.0081,0.0636]$,
includes zero despite a positive mean difference. Thus, the source-level
analysis supports the Macro-F1 improvement against both encoders while
showing greater variation in the balanced-accuracy comparison with
SkateFormer. Resampling complete recordings preserves the dependence among
windows within each source.

Additional probability-based metrics give NLL/Brier scores of
$0.3556/0.1709$ for \method, $0.4943/0.2073$ for SkateFormer, and
$0.5297/0.2173$ for CTR-GCN. Negative log-likelihood and the Brier score both
favor the fused predictions. These post-selection evaluations complement ECE
by assessing the full probability distribution and were not used to select
$\alpha$.

\subsection{Qualitative Complementarity Analysis}

\begin{figure}[t]
  \centering
  \includegraphics[width=0.48\textwidth]{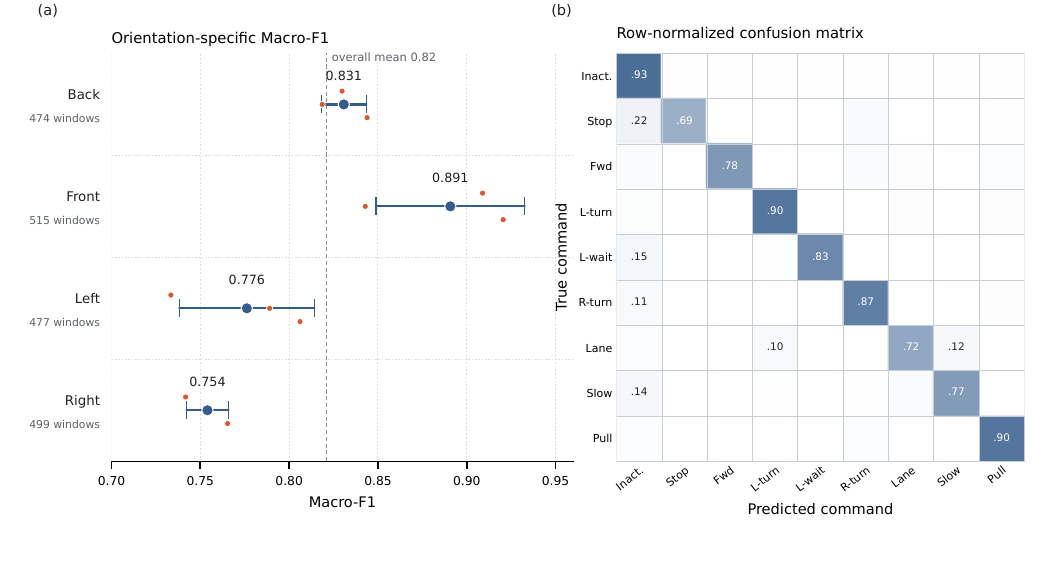}
  \caption{View-dependent performance and class confusion.
(a) Macro-F1 across viewing directions (mean $\pm$ standard deviation over
three seeds); dots show individual seeds and the dashed line denotes the
overall mean. (b) Three-seed mean row-normalized confusion matrix, with
diagonal entries and off-diagonal values $\geq 0.08$ annotated.}
  \label{fig:diagnostics}
\end{figure}

The selected examples comprise two cases in which only one branch is correct and one case in which both branches agree. Their pose sequences and seed-1 predictions are reported in Fig.~\ref{fig:qualitative} and Table~\ref{tab:qualitative}. In panel (a), CTR-GCN
correctly predicts Forward with confidence $1.00$, while SkateFormer favors
Left turn with confidence $0.55$. Fusion retains Forward with confidence
$0.72$. Panel (b) shows the complementary case: SkateFormer correctly
predicts Pull over with confidence $0.98$, while CTR-GCN predicts Right turn.
The fused prediction is Pull over with confidence $0.71$.

Panel (c) illustrates cross-branch agreement. Both encoders correctly
predict Left turn, with confidences $1.00$ and $0.99$, and fusion preserves the
decision at $0.99$. Together, the three cases show that the fused classifier
retains the correct branch decision under disagreement while preserving a
clear decision when the spatial and temporal encoders agree.

\subsection{Ablation Study}

The validation-selected scalar mixture achieves the highest Macro-F1 and balanced accuracy among the evaluated fusion variants in Table~\ref{tab:ablation}. Temperature calibration remains close, with Macro-F1 and balanced accuracy lower by approximately $0.0004$ and $0.0011$, respectively. Groupwise fusion is lower by $0.0143$ Macro-F1 and $0.0103$ balanced accuracy, while the random-weight control is lower by $0.0100$ and $0.0083$, respectively.

\begin{table}[t]
  \centering
  \caption{Score-level fusion ablations on the nine-class test protocol.
  Scores are shown at higher precision than in Table~\ref{tab:main} to resolve
  the small differences between variants.}
  \label{tab:ablation}
  \small
  \setlength{\tabcolsep}{4.8pt}
  \begin{tabular}{lcc}
    \toprule
    Variant & Macro-F1 $\uparrow$ & Bal. Acc. $\uparrow$ \\
    \midrule
    \textbf{\method} & \textbf{.821049} & \textbf{.821638} \\
    Temperature calibration & .820686 & .820537 \\
    Groupwise fusion & .806748 & .811305 \\
    Random-weight control & .811037 & .813306 \\
    \bottomrule
  \end{tabular}
\end{table}

The controls examine posterior concentration, class-dependent weighting,
and coefficient selection. Among these variants, the validation-selected
scalar mixture gives the highest Macro-F1 and balanced accuracy. The
comparison with groupwise fusion favors a shared coefficient for the
available validation data, while the random-weight result supports selecting
the mixture from validation evidence. The calibration variant remains close
to the original mixture, suggesting that its principal recognition behavior
is retained after probability rescaling. These results favor the validation-selected scalar mixture
for the current branch pair and data split.

\subsection{Detailed Analysis of Recognition Results}

Recognition performance varies across officer orientations and gesture classes, as summarized in Fig.~\ref{fig:diagnostics}. Macro-F1 is $0.891$ for Front, $0.831$ for Back, $0.776$ for
Left, and $0.754$ for Right. The front and back views therefore have higher
mean scores than the lateral views. Class-wise F1 is $0.935$ for Inactive,
$0.732$ for Lane change, $0.765$ for Stop, and $0.783$ for Slow down. The
confusion matrix provides a complementary recall-based view: Stop,
Left-turn waiting, and Slow down are assigned to Inactive at rates of
$0.22$, $0.15$, and $0.14$, respectively. Lane change is most often confused
with Slow down ($0.12$) and Left turn ($0.10$).

These patterns locate the remaining errors among inactive transitions and
related commands. The orientation analysis further identifies lateral views
as an important source of variation within the evaluation set. Considered
alongside the overall and class-balanced gains, the results indicate that
posterior fusion improves command recognition while retaining sensitivity
to viewpoint and temporal ambiguity. The recording-level comparisons and
qualitative cases explain how complementary branch evidence contributes to
this improvement at different levels of analysis.

\FloatBarrier
\section{Conclusion}

We presented \method, a graph--transformer framework that integrates
independently learned skeletal representations through validation-selected
posterior fusion. On the nine-class CTPGesture v2 protocol, it achieves
$88.82\%$ accuracy and $82.10\%$ Macro-F1, improving on the strongest
individual branch for each metric by $2.15$ and approximately $3.8$ percentage
points, respectively. Pairwise decision analysis, recording-level comparisons, and
fusion ablations show how complementary graph and attention evidence
contributes to class-balanced recognition. Future work will investigate
causal temporal windows, viewpoint-robust representations, and target-hardware
evaluation.

\bibliographystyle{IEEEtran}
\bibliography{references}

\end{document}